# Latent Space is Feature Space
## Regularization Term for GANs Training on Limited Dataset

Pengwei Wang

*School of Software, Northwestern Polytechnical University, Xi'an, 710129, China*

*Corresponding author. Email: *penway@mail.nwpu.edu.cn*

**ABSTRACT**

Generative Adversarial Networks (GAN) is currently widely used as an unsupervised image generation method. Current state-of-the-art GANs can generate photorealistic images with high resolution. However, a large amount of data is required, or the model would be prone to generate images with similar patterns (mode collapse) and bad quality. I proposed an additional structure and loss function for GANs called LFM, trained to maximize the feature diversity between the different dimensions of the latent space to avoid mode collapse without affecting the image quality. Orthogonal latent vector pairs are created, and feature vector pairs extracted by discriminator are examined by dot product, with which discriminator and generator are in a novel adversarial relationship. In experiments, this system has been built upon DCGAN and proved to have improvement on Fréchet Inception Distance (FID) training from scratch on CelebA and MetFace Dataset. This system requires mild extra performance and can work with data augmentation methods. The code is available on github.com/penway/LFM.

## 1. INTRODUCTION

The adversarial networks framework is an adversarial game between two deep neural networks, where one network tries to generate synthetic data and the other tries to diminish the fake from the real [1]. Deep convolutional generative adversarial network (DCGAN) is then proposed to use convolution and transpose convolution to keep spatial information of images [2]. All state-of-the-art GANs are based on this simple yet effective structure[3-7].

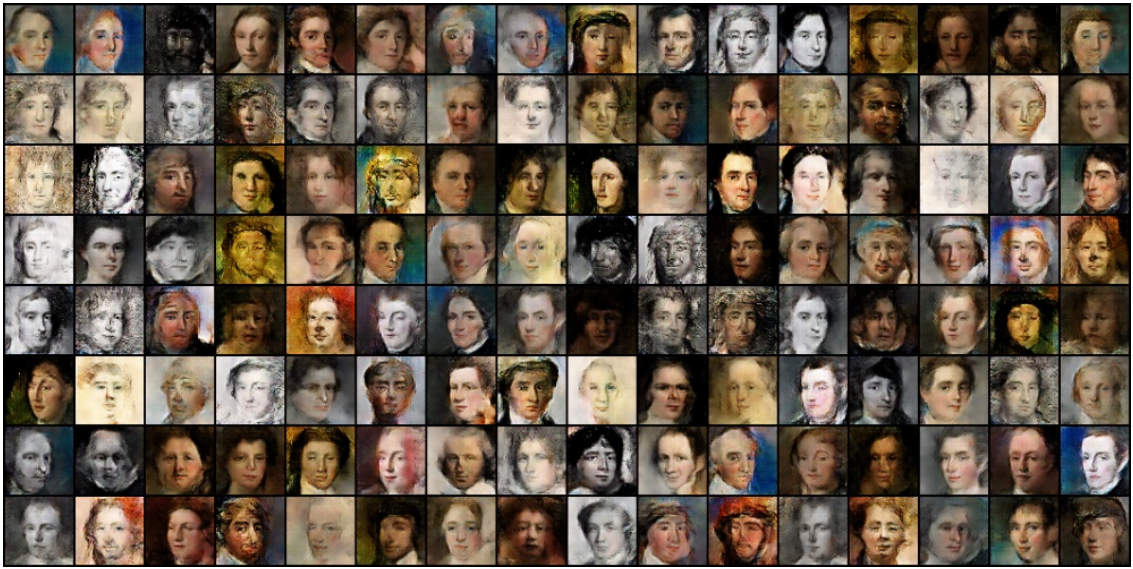

Figure 1. Image generated by DCGAN-LFM trained from scratch on MetFace Dataset (1336 images) [Owner-draw]

The authenticity of GANs largely relies on the size of the dataset; with fewer images comes poorer quality. When the number of training images is small enough, GANs deteriorate in various behaviours, for DCGAN is mode collapse, and WGAN stops generating meaningful images [3].

In this paper, I propose *Latent Feature Maximization* strategy as a regularization term for the loss function of GANs, which can slow down the procedure of mode collapse and reach a lower FID score before that. The result of training from scratch on MetFace dataset is shown above in Figure 1.

This paper is in the following structure: 2. Related Work, 3. Methodology, 4. Experiment results, 5. Conclusion.

## 2. RELATED WORK

### *2.1. Generative Adversarial Networks*

The Generative Adversarial Networks was first proposed by Goodfellow et al, using an adversarial setting to train the generative model [1]. It is theoretically proved that this minimax game has a global optimum for $P(G) = P(data)$. Despite many advantages of the GAN framework, disadvantages are mode collapse and the hardship of keeping a balance between $D$ and $G$, and this system will not converge but will finally deteriorate with training going on. Mode collapse means that $G$ has output the same $x$ from different $z$, meaning lacking diversity. The balance problem means that neither $D$ nor $G$ should be too strong to overshadow each other, or this minimax game would diverge.

While utilizing convolutional layers to scale up GAN is a natural thought, attempts before the paper from Radford et al. were not successful. With many detailed implementation advice and settings, the paper also explored the internal properties of the networks [2]. By testing on linear changing latent space, the model is proved to have learned feature representations instead of memorizing the examples. Images can be interpolated with vector arithmetic in the latent space, resulting in the corresponding feature addition or subtraction.

### *2.2. Fréchet Inception Distance*

The Fréchet Inception Distance is proposed by as a measure the generate image quality and diversity [8]. Only with both quality and diversity high can the result be small. FID is the Fréchet Distance between feature vectors encoded by the backbone encoder of the Inception v3 model.

### *2.3. Methodologies for GAN training on small datasets*

**TransferGAN**: Wang et al. proposed utilizing pre-trained models to train on a new dataset [9]. They explored the relationship between dataset size and if pre-trained, showing that the pre-trained model can always achieve a better score; with a smaller new dataset came more significant improvement.

**MineGAN**: This paper proposed a novel component for GANs called *Miner* to estimate the proper distribution of the latent space for the generator [10]. Such components can be used in one or multiple pre-trained models. The base model used in this paper is WGAN-GP [4]. The result of MineGAN outperforms TranferGAN. Moreover, this novel architecture can utilize pre-trained models from multiple domains.

**ADA**: After a Data Augmentation methods Balanced Consistency Regularization proposed by Zhao et al., Karras et al.proposed a data augmentation algorithm called *Adaptive Data Augmentation*, in which an augmentation pipeline a designed to reduce overfitting and an adaptive algorithm a designed to change to the possibility to augment data during training dynamically [11,12].

**LC regularization**: Tseng et al.proposed a regularizing method for GAN [13]. Together with ADA, it is proven to achieve state-of-the-art performance even with limited data. This loss function is theoretically connected with LeCam Divergence, which is also a member of the f-divergence family and proved to be more robust when facing small datasets compared to other divergence functions.

## 3. METHODOLOGY

### 3.1. Loss function of GAN

DCGAN follow the basic structure of GAN proposed by Goodfellow et al. The network consists of one generator $G$ and one discriminator D. G map latent code $z$ into fake image $\hat{x}=G(z)$, and $D$ map image $x$ into class $c$. The target of D is trying to maximize the correct rate of itself discriminating fake images from real, while the generator tries to minimize it. In other words, D is maximizing $V_D$ and $G$ is minimizing $L_G$.

$$V_D = \mathbb{E}[log(D(x))] + \mathbb{E}[log(1 - D(\hat{x}))] \quad (1)$$

$$V_G = E[log(1 - D(\hat{x}))] \quad (2)$$

### 3.2. Analysis of the mode collapse

Even with a small dataset, the first few iterations can also show the same tendency of FID with larger datasets, and no mode collapse is happening in this phase. After certain training iterations, generated images gradually collapse to similar patterns, with better quality but diversity close to none. The current strategy is to utilize the best model before mode collapse happens. With a rather small dataset, DCGAN can show instability and step into mode collapse in a rapid manner. Generated images can jump between meaningless noise and mode collapse images during training.

Mode collapse means that despite that given noise for the generator being always random, a generator with mode collapse maps all different noises into several fixed modes. As a result, a regularization term $LFM$ is introduced based on the following assumption.

Assumption: The latent space is the same as the feature space. Given a latent space fit for the network capacity, the latent space for $G$ to encode can also be interpreted as feature space that the feature extractor $D_F$ decoded.

With the assumption above, no matter the choice of origin and scale in the space, orthogonal vectors remain the dot product equals 0. In the following equations, $D$ is split into backbone feature extractor $D_F$ and classifier $D_C$, so $D(x) = D_C(D_F(x))$.

$$\begin{aligned} f &= D_F(G(z)) \\ f_1^T f_2 &= 0 \text{ such that } z_1^T z_2 = 0 \end{aligned} \quad (3)$$

### 3.3. Latent Feature Maximization

To evaluate the performance of whether orthogonal vectors in latent space are properly mapped into orthogonal vectors in feature space, pairs of orthogonal vectors are generated for the generator $G$, and the discriminator has a model $D_F$ to encode features, the classifier $D_C$ in DCGAN is also kept and shares the same parameters with $D_F$. To form an adversarial loss function, the feature extractor has one trainable convolution layer to maximize the loss function. As a result, actually $f = F(D_F(x))$ is used to approximate $f = D_F(x)$. The structure is shown in Figure 2.

One thought would be that the discriminator only optimizes classifier loss and has a natural feature vector to optimize the generator. This turned out to be not useful and only had similar results without the regularization term for $G$. The experiment will be shown in 4.3.

$$\begin{aligned} &\text{latent pairs: } z_{i+} \cdot z_{i-} = 0 & & R_{LFM} = \mathbb{E}[f_{i+} \cdot f_{i-}] \\ &\text{Fake: } \hat{x}_{ij} = G(z_{ij}) & & \max_D L_D, L_D = V_D + \lambda_D R_{LFM} \\ &\text{Feature: } f_{ij} = D_F(\hat{x}_{ij}) & & \min_G L_G, L_G = V_G + \lambda_G R_{LFM} \end{aligned} \quad (4)$$

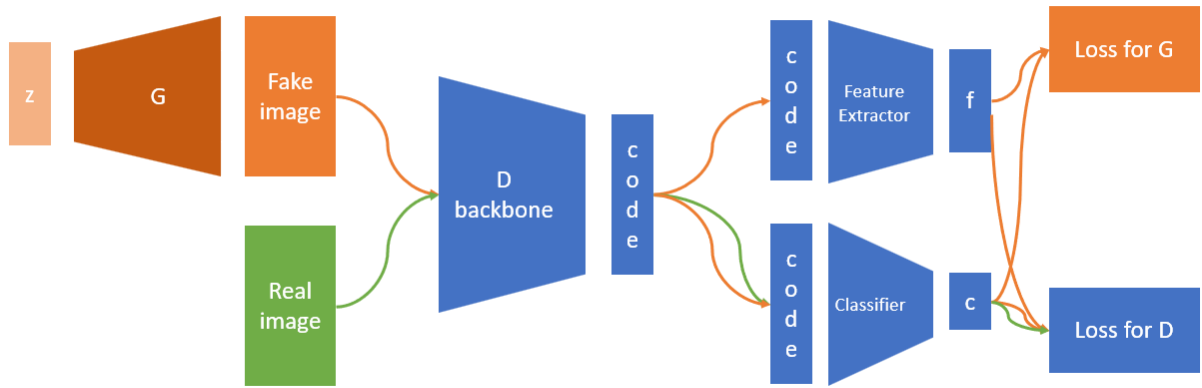

**Figure 2. Structure of LFM [Owner-draw]**

### 3.4. Generate Orthogonal Vector Pairs

Noise for DCGAN is under Gaussian distribution. As a result, an algorithm is needed to generate orthogonal vectors for a given vector without largely away from the prior distribution. A very simple Algorithm 1 to generate vector counterparts is currently used.

**Algorithm 1:** Generate Orthogonal Vector Pairs
```
Data: None
Result: Noise List shape like (batchsize, z_dim)
y ← 1;
X ← x;
N ← n;
While count < batchsize/2 do
    n1 ← RandomVector;
    n2 ← RandomVector;
    dr ← DotProduct(n1[0:-1], n2[0:-1])
    n2[-1] ← -dr/n1[-1]
    if abs(n2[z_dim-1] <= 1 then
        Add Pairs To Noise;
        count ← count + 1;
    end
end
```

Note: there is an absolute function in the if statement.

Experiments are also done with $n2[z_dim - 1] \leq 1$, leaving about 23% of the noise out of the absolute value of 1 in the last element, which resulted in a better outcome in small datasets, even without LFM. The related study will be in 6.3. The following sections will use the algorithm with an absolute function.

### 3.5. Implementation Details

#### 3.5.1. Network Details

DCGAN in this paper follows the structure and hyperparameters in and instructions from [2, 14, 15]. Network detail is in Table 2.

Weights for convolutional layers are initialized to mean 0 and standard deviation 0.02. Weights for batch normalization are initialized to mean one and standard deviation 0.02.

Table 2. Network Details [Owner-draw]

| $D_C$ | $D_F$ | $G$ |
|---|---|---|
| Conv 4x4 (3, 64) | | TransConv 4x4 (100, 512) |
| | | BatchNorm |
| LeakyReLU 0.2 | | ReLU |
| Conv 4x4 (64, 128) | | TransConv 4x4 (512, 256) |
| BatchNorm | | BatchNorm |
| LeakyReLU 0.2 | | ReLU |
| Conv 4x4 (128, 256) | | TransConv 4x4 (256, 128) |
| BatchNorm | | BatchNorm |
| LeakyReLU 0.2 | | ReLU |
| Conv 4x4 (256, 512) | | TransConv 4x4 (128, 64) |
| BatchNorm | | BatchNorm |
| LeakyReLU 0.2 | | ReLU |
| Conv 4x4 (512, 1) | Conv 4x4 (512, 100) | TransConv 4x4 (64, 3) |
| Sigmoid | Tanh | Tanh |

### 3.5.2. Loss Function

Loss functions for networks need to be minimised. As a result, the loss function for $G$ is just $R_G = \mathbb{E}[f_{i+} \cdot f_{i-}]/2$, while for $D$, currently a naive one is used, shown below as $100 - R_G$. 100 is the maximum possible value for $R_G$ when all elements in feature vector are 1.

Better loss function and interpretation of LFM might exist with entropy or in other forms. Other loss functions are not tested.

Code in PyTorch is shown below.

**Algorithm 2: LFM Regularization Loss Funtion**

```python
class FeatureLoss(nn.Module):
    def __init__(self):
        super(FeatureLoss, self).__init__()

    def forward(self, x, isDisc):
        dot_product = 0
        hb = int(len(x) / 2)
        for i in range(hb):
            dot_product += torch.dot(x[i].view(-1), x[i+hb].view(-1))
        loss_base = torch.abs(dot_product / hb / 2)

        if isDisc:
            return 100 - loss_base
        else:
            return loss_base
```

## 4. EXPERIMENTAL RESULT

### 4.1. Basic Settings

Dataset. CelebA and MetFace. For CelebA, images are centre cropped to a square and resized to 64x64 [12, [16]. Dataset is made by taking the first few photos, ranging from 0.25% (512 images) to 10% of the full dataset (202,59 images). MetFace dataset originally contains 1366 high-quality images; they are also resized to 64x64.

Evaluation metric. Fréchet Inception Distance is used [8]. In practice, the PyTorch-fid package is used, which is reported to have a tiny difference from the original implementation [17]. All baseline statistics are from the full

CelebA dataset calculated by PyTorch-fid, while generated image set only contains 128 images for each calculation. Despite specially structured training noise, validation noise is fully random to keep the metric fair. One note I have discovered is that the pre-calculated statistics in the original implementation for CelebA are for face only (zoomed). As a result not suitable for the dataset with full head shown [8].

Setup. Both DCGAN and DCGAN-LFM are built on PyTorch from scratch, strictly following the original implementation of DCGAN, including hyperparameters, since it is indeed the best setting.

Baseline. Only DCGAN is used in this paper.

**Table 1. Min FID of DCGAN and DCGAN-LFM in three different dataset sizes. [Owner-draw]**

|  | **CelebA** [16] | | | | **MetFace** [12] |
| --- | --- | --- | --- | --- | --- |
|  | 512 images | 1012 images | 2024 images | 20k |  |
| **DCGAN** [2] | 158.48 | 148.44 | 119.57 | **85.04** | 138.06 |
| **DCGAN-LFM** (mine) | **152.01** | **125.72** | **108.49** | 85.98 | **119.30** |

### 4.2. Result

LMF actually cannot avoid the coming of the mode collapse phase but can delay its arrival and reach a better FID score before that. While DCGAN becomes unstable with a dataset with less than 2k images, LFM can still offer stable training.

As shown in Table 1, DCGAN with LFM can achieve a lower minimum FID when the dataset is small while have a tendency to decrease performance with larger dataset.

As is shown in Figure 3, for the dataset with 512 images, DCGAN can achieve FID 174.44 before mode collapse and 158,48 after it. While with LFM, FID 152.01 can be achieved before mode collapse, and the FID curve becomes more stable. The situation is similar for the dataset with 1012 images.

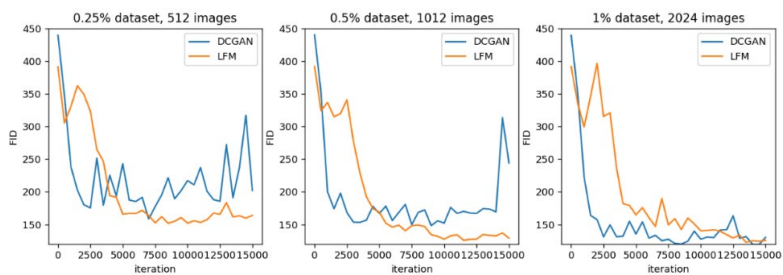

**Figure 3. FID curve of DCGAN and DCGAN-LFM training on three different datasets. [Owner-draw]**

In the rightmost figure in Figure 3, LFM do not achieve the minimum FID within 15000 iterations, while achieve it at 24000 iterations. As is shown in Figure 4, training on different dataset all shown similar pattern. However, training with LFM on 200k dataset did not show clear better result, which is a unsolved problem.

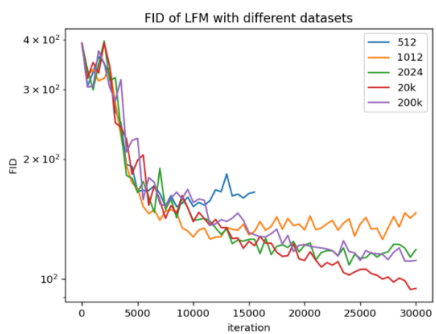

**Figure 4. FID curves of DCGAN-LFM training by four different sizes of datasets. Training on the dataset with 512 images was already stopped early for showing mode collapse. [Owner-draw]**

### *4.3. Ablation Study: LFM loss on G only*

I have also tried to use naturally generated feature only with one layer of tanh activation, without a additional trainable fully connection layer. The structure of *LFM loss on G only* is shown below in Figure 5.

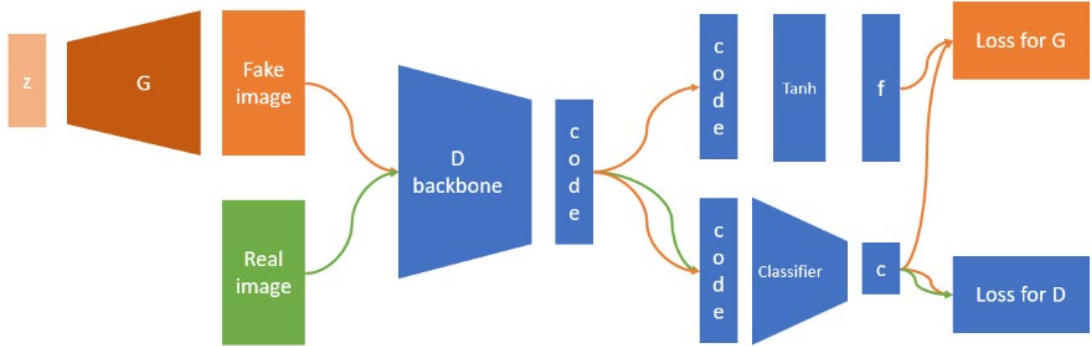

**Figure 5. Structure of LFM loss on G only [Owner-draw]**

The experiment is run on the dataset with 1012 images and shows almost the same FID curve compared to the original DCGAN. The result graph is shown in Figure 6. This strategy might have slight improvement a the beginning stage before the mode collapse, but have a similar min FID and does not show any promise to keep on training.

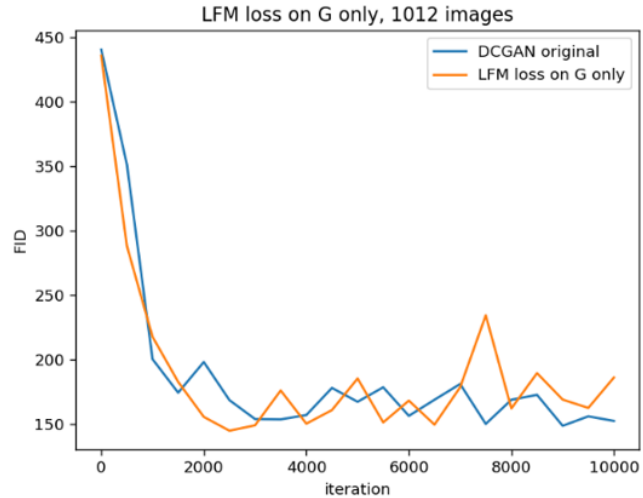

**Figure 6. Comparing DCGAN and LFM loss on G only. [Owner-draw]**

## 5. CONCLUSION

In this paper, I proposed *Latent Feature Maximization (LFM),* one novel regularization term for GANs. It pushed the generator to map orthogonal latent vector pairs to orthogonal feature vectors pairs in discriminator in an adversarial manner. In practice, orthogonal feature vector pairs are generated for training, and after being encoded into feature vectors by the discriminator, the loss is calculated accordingly.

The proposed LFM regularization is proved to have an improvement to DCGAN when the dataset size is small (less than 20k images in CelebA and full dataset MetFace). It can offer more stable training and lower minimum FID, and thereotically can work with other GAN backbones and data augmentation methods. In future work, I plan to 1) experiment on more datasets and backbones; 2) examine latent space intrinsic after applying LFM.